\documentclass[10pt,twocolumn,letterpaper]{article}

\usepackage{cvpr}              
\usepackage[accsupp]{axessibility}  

\usepackage{amsmath,amsfonts,bm}

\def\eqref#1{equation~\ref{#1}}

\def\1{\bm{1}}

\DeclareMathAlphabet{\mathsfit}{\encodingdefault}{\sfdefault}{m}{sl}
\SetMathAlphabet{\mathsfit}{bold}{\encodingdefault}{\sfdefault}{bx}{n}

\usepackage{algorithm}
\usepackage{algpseudocode}

\algrenewcomment[1]{\(\triangleright\) #1}
\algnewcommand{\LineComment}[1]{\State \(\triangleright\) #1}

\usepackage{colortbl,xcolor}
\usepackage{adjustbox}

\definecolor{cvprblue}{rgb}{0.21,0.49,0.74}
\usepackage[pagebackref,breaklinks,colorlinks,allcolors=cvprblue]{hyperref}

\newcommand{\Ours}{\textsc{Act2See}\xspace}

\def\paperID{3768} 
\def\confName{CVPR}
\def\confYear{2026}
\newcommand{\think}[1]{\textcolor{blue}{\texttt{<think>}} #1 \textcolor{blue}{\texttt{</think>}}}
\newcommand{\retrieve}[1]{\textcolor{cyan}{\texttt{<retrieve>}} #1 \textcolor{cyan}{\texttt{</retrieve>}}}
\newcommand{\generate}[1]{\textcolor{brown}{\texttt{<generate>}} #1 \textcolor{brown}{\texttt{</generate>}}}
\newcommand{\answer}[1]{\textcolor{red}{\texttt{<answer>}} #1 \textcolor{red}{\texttt{</answer>}}}
\newcommand{\frames}[1]{\textcolor{magenta}{\texttt{<frame>}} #1 \textcolor{magenta}{\texttt{</frame>}}}

\title{\Ours: Emergent Active Visual Perception for Video Reasoning}

\author{Martin Q. Ma$^{1}$
\and 
Yuxiao Qu$^{1}$
\and
Aditya Agrawal$^{1}$
\and
Willis Guo$^{1}$
\and
Paul Pu Liang$^{2}$
\and
Ruslan Salakhutdinov$^{1}$
\and
Louis-Philippe Morency$^{1}$
\and
\\
$^1$Carnegie Mellon University \hspace{5em} 
$^2$MIT
}

\begin{document}
\maketitle
\begin{abstract}
Vision-Language Models (VLMs) typically rely on static initial frames for video reasoning, restricting their ability to incorporate essential dynamic information as the reasoning process evolves. Existing methods that augment Chain-of-Thought (CoT) with additional frame information often exhibit suboptimal CoT quality and lack the crucial ability to synthesize visual information for hypothetical or counterfactual scenarios. We introduce Act-to-See (\Ours), a novel framework that enables active visual perception by empowering VLMs to actively interleave video frames within text CoTs. \Ours is developed via Supervised Fine-Tuning (SFT) on a high-quality dataset of reasoning traces generated by a frontier VLM. These traces integrate active calls to either retrieve existing frames or generate new ones, and are rigorously verified against human-annotated CoTs to ensure quality. This approach cultivates an emergent capability: at inference time, the model actively determines when to search for or synthesize the necessary visual evidence. \Ours establishes new state-of-the-art results on challenging benchmarks, including VideoEspresso and ViTIB, and outperforms comparable or larger models on Video-MME, EgoNormia, and VCR-Bench, demonstrating an advancement in enabling VLMs with active visual perception for video reasoning. Code: \url{https://github.com/martinmamql/act2see}.
\end{abstract}    
\section{Introduction}
\label{sec:intro}
Chain-of-Thought (CoT) prompting \citep{wei2022chain} has been integrated into Vision-Language Models (VLMs), and such a paradigm has been shown to enhance VLM reasoning capabilities on complex video tasks \citep{hurst2024gpt, comanici2025gemini, bai2025qwen2, chen2024expanding, zhang2025videollama, bai2025qwen3}. The standard VLMs take a set of static initial video frames as input, or a video as input and sample at a fixed frame-per-second (fps). However, unlike image reasoning, video reasoning often involves nuanced visual details in spatiotemporal dynamics \citep{ge2025famemind, zhang2025rewatch, han2025videoespresso, wang2024videocot, nagrani2025minerva}, which may be absent from the initial frames and require additional information. Static input frames can prevent the VLMs from actively collecting new evidence as reasoning evolves.

Recent efforts have aimed to improve video reasoning with CoT by adding video frame information, such as pre-determined keyframes, visual tool calls, localized temporal captioning, or key frame IDs to the CoTs \citep{zhang2025vitcot, hu2025cos,ge2025famemind, zhang2025rewatch, ghazanfari2025chain}. However, there are two challenges the existing methods face. First, \textbf{\textit{CoT quality}} can be suboptimal: frame-information-augmented CoTs are often derived from VLM-generated ground truth CoTs \citep{feng2025video, han2025videoespresso, ge2025famemind}. However, as shown in \citet{zhang2025rewatch}, automated CoTs generated without human verification can sometimes be suboptimal and hinder the performance of the model. Second, it is crucial to \textbf{\textit{generate hypothetical scenarios}}. Based on rich knowledge of the world, VLMs should have the ability to produce hypothetical scenarios, generate potential outcomes, and offer causally grounded
responses \citep{foss2025causalvqa, li2024eyes}. However, existing video reasoning methods cannot synthesize new visual frames inside CoTs \citep{ge2025famemind, zhang2025rewatch, ghazanfari2025chain}. Recent video reasoning benchmarks \citep{qi2025vcr, foss2025causalvqa, nagrani2025minerva, rezaei2025egonormia} have started to focus on evaluation with real-world counterfactual or hypothetical questions where the requested scenarios are not present: for example, asking what will happen to an object if the temporal order in the original video had switched, and existing methods cannot ``imagine'' such scenarios visually during reasoning. These two challenges prevent video reasoning models from achieving better reasoning capabilities. 

\begin{figure*}[th]
	\centering
  \includegraphics[width=1\linewidth]{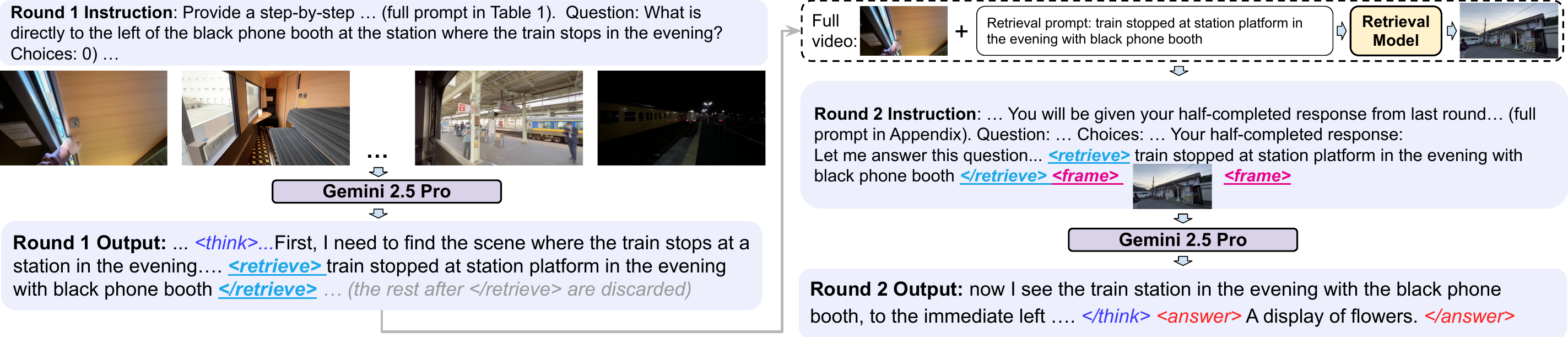}
  \caption{The two-round construction of the interleaved video-text CoT dataset. In the first round, Gemini 2.5 Pro is prompted to perform video reasoning, and simultaneously output a query to retrieve or generate any additional video frame when necessary. The retrieval or generation is done by offline models. In the second round, Gemini 2.5 Pro is asked to complete the reasoning based on the round 1 response (up to the point of the retrieval or generation query) and the new frame. The final reasoning CoT will be the concatenation of round 1 and round 2 outputs, which consists of the text reasoning interleaved with the retrieved or generated frame.}
  \label{fig:dataset}
\end{figure*}

We present \textbf{Act-to-See (\Ours)}, a new framework that employs Supervised Fine-Tuning (SFT) to enable active visual perception that addresses the above challenges. By active visual perception \citep{bajcsy1988active,zhang2025rewatch, ge2025famemind, su2025openthinkimg, zheng2025deepeyes}, we mean actively deciding when and how to gather new visual information during video reasoning. To enable the capability of active visual evidence acquisition by searching existing videos or generating hypothetical scenarios, we prompt a frontier model (Gemini 2.5 Pro \citep{comanici2025gemini}) to output CoTs with specialized tool call syntax (i.e., retrieval or generation tokens and associated queries). The model actively inserts retrieval or generation tokens and corresponding queries whenever it determines the existing visual information is insufficient. Offline retrieval or generation models are leveraged to gather the new frame. To ensure quality, we leverage video reasoning datasets with human-annotated ground-truth CoTs: MINERVA \citep{nagrani2025minerva}, CausalVQA \citep{foss2025causalvqa}, and Social Genome \citep{mathur2025social}, and use the human annotated ground-truths as gold standard to filter out CoTs whose similarity scores with the corresponding ground-truths are low. We construct this high-quality video reasoning SFT dataset with almost half ($47\%$) of the traces containing retrieved or generated frames inside the CoTs. We use the resulting dataset to fine-tune a base model, Qwen3-VL-8B-Thinking \citep{bai2025qwen3}. 

\Ours shows significant performance improvements on five video reasoning benchmarks: Video-MME \citep{fu2025video}, VideoEspresso \citep{han2025videoespresso}, EgoNormia \citep{rezaei2025egonormia}, VCR-Bench \citep{qi2025vcr}, and ViTIB \citep{zhang2025vitcot}, encompassing spatiotemporal reasoning, physical-social reasoning, complex perception, logical reasoning (including counterfactual), and video-text interleaving verification. It outperforms all standard VLMs of similar size such as InternVL2.5-8B \citep{chen2024expanding}, Video-LLaMA3-7B \citep{zhang2025videollama}, Qwen3-VL-8B-Thinking \citep{bai2025qwen3}, and Video-R1 \citep{feng2025video}. \Ours also outperforms most video-text interleaved CoT methods, such as Chain-of-Shot \citep{hu2025cos}, FrameMind \citep{ge2025famemind}, and ReWatch-R1 \citep{zhang2025rewatch}. More importantly, we observe the emergent capability of the model in retrieving or generating new visual frames inside the CoT at inference time (Figure \ref{fig:inference_examples}), enabling VLMs to actively decide when to search or synthesize new visual information during reasoning.
\section{Method}

Our goal is to enable a VLM with active visual perception capabilities of gathering new visual information inside the CoT. To achieve this goal, we design an SFT dataset with interleaved video frame and text in the CoT, by prompting a frontier model to include frame retrieval or generation calls during reasoning. In this section, we lay out the overview of the dataset construction algorithm as well as the supervised-finetuning (SFT) training objective, and provide the details of the construction in Sect. \ref{sec:dataset_description}. We include an overview of the data construction process in Figure \ref{fig:dataset}.

\subsection{Creating interleaved CoTs for SFT}
We develop a pipeline to automatically construct interleaved video-text CoTs using an off-the-shelf multimodel model together with our instruction template (Table \ref{tab:instruction}). The goal of this pipeline is simply to generate explicit reasoning traces and visual operations from a strong VLM, not to distill or imitate any particular model.
As shown in Table \ref{tab:instruction}, the VLM is instructed to structure its reasoning within specialized tokens, 
where the \retrieve{$\dots$} or \generate{$\dots$} are the tool call tokens, and the \textit{retrieval prompt} or the \textit{generation prompt} inside is provided by the VLM simultaneously along with the tool call (e.g., \generate{\textit{a bison baby chased by a wolf pack}}). 

Once the VLM generates the response in the first pass, if any tool call tokens are detected ({\textcolor{cyan}{\texttt{<retrieve>}} or {\textcolor{brown}{\texttt{<generate>}}), the \textit{retrieval prompt} or \textit{the generation prompt} will be extracted, and a separate frame retrieval or generation tool will be called and the resulting retrieved or generated frame will be returned. Note that to increase generation quality, we first retrieve a frame based on the generation query, and then use the retrieved frame and the generation query for conditional image-to-image generation. 

To obtain complete, cohesive CoT with the newly gathered frame, everything from the VLM's first response, up to the end of \retrieve{\textit{retrieval prompt}} or \generate{\textit{generation prompt}}, will be retained, and the new visual frames will be added to it. This process is shown in Algorithm \ref{alg:interleaved} (following \citet{zhang2025rewatch} format), which applies to both gathering the SFT data and inference of the model. The retrieved frame is different from the initial frames as the retrieval model searches the video at a higher sampling rate. Importantly, the ground-truth CoTs are not used during generation; they are leveraged later for quality control, as explained in Sect. \ref{sec:data_quality_control}. This is due to ground-truth CoTs as input significantly lowering the chances of calling retrieval or generation tools, as shown in Sect. \ref{sec:prompt_with_ground_truths}. Also, if no retrieval or generation query is detected, such text-only traces will also be considered, creating a blended set of interleaved video-text CoTs with text-only CoTs. 

This procedure produces a warm-up dataset with explicit, grounded multimodal reasoning steps. Importantly, any model capable of following the instruction template can be used in this pipeline. For implementation simplicity, we use Qwen-3-VL-8B-Thinking in our experiments, though the method is model-agnostic.

\subsection{Supervised Fine-Tuning with Interleaved CoT}
Given the input video $\bm{V}$ represented as a sequence of sampled frames $V = \{v_1, v_2, ..., v_T\}$, the input textual question of the video as a sequence of tokens $Q = \{q_1, q_2, ..., q_K\}$, the input answer choices in free-form texts as a sequence of tokens $A = \{a_1, a_2, ..., a_M\}$, the ground truth answer along with the CoT from the new dataset as a sequence of tokens $Y = \{y_1, y_2, ..., y_L\}$, a separate base model we perform SFT on (which is a different model from Gemini 2.5 Pro where we collect CoTs): $\pi_\theta$, and the video QA dataset $\mathcal{D}$: $\mathcal{D} = \{(V^{(i)}, Q^{(i)}, A^{(i)})\}_{i=1}^N$ where each entry corresponds to a single triplet of one video and one question and the corresponding answers.
The loss function of the SFT will be the following:
\begin{gather}
J(\theta) = - \frac{1}{N} \sum_{i=1}^{N} \sum_{t=1}^{L^{(i)}} \log P\big(y_t^{(i)} | V^{(i)}, Q^{(i)}, y_{<t}^{(i)}; \theta\big)
\end{gather}
We compute the token-level losses over the whole rollout, including sequences of retrieval or generation queries, only excluding retrieved or generated frames. 

\paragraph{Inference.} After training the base model with SFT, to perform video reasoning at inference time, we also follow Algorithm \ref{alg:interleaved}: the trained model outputs a first-round response, where retrieval or generation by offline models will be executed if retrieval or generation queries are detected, and then the new frame will be added to the original response as a pretext for completion in the second round. If no retrieval or generation query is detected, the reasoning will be in text only. The final answer will be extracted and compared with the ground-truth target.
\begin{table*}[h]
    \centering
    \begin{tabular}{p{13cm}}
        \hline
        Provide a step-by-step answer for the question below. You will be provided a set of initial video frames. 
        Your task is to generate the reasoning traces, which include texts and images, to correctly answer the question.
        You must conduct reasoning inside \think{and}. When reasoning, if you need any visual
        knowledge, you can call a frame retrieval module by \retrieve{\textit{retrieval prompt}},
        or a frame generation module by \generate{\textit{generation prompt}}, and either will
        return the video frame between \frames{and}. The retrieval or generation prompt is the text description of the frame you want to retrieve/generate and you should provide it. ... (an example omitted here).
        Finally, give the answer in free-form text in the end, with \answer{and} tags. \\
        \hline
    \end{tabular}
    \caption{Template for \Ours. The specific questions and videos are added to the end of the template.}\label{tab:instruction}
\end{table*}

\begin{algorithm}[t]
\caption{Algorithms for \Ours for both data generation and inference.}
\label{alg:interleaved}
\begin{algorithmic}[1]
\Require Input initial frames $V$, question $Q$, answer choices $A$, full video $\bm{V}$, VLM $f$, retrieval model $h_r$, and generation model $h_g$ \
\Ensure Final full response $Y$

\State Initialize final full response \( Y \gets \emptyset \)
\LineComment{\textbf{Round 1: add a new frame if necessary}}
\While{True}
\State Generate response token \( y_t \sim f(\cdot \mid V, Q, A, s_{<t}) \)
\State Append \( y_t \) to full sequence \( Y \gets Y + y_t \)
\If{\( y_t \) in [\textcolor{cyan}{\texttt{</retrieve>}}, \textcolor{brown}{\texttt{</generate>}}, \textcolor{red}{\texttt{</answer>}}, \texttt{<eos>}]}
    break   
\EndIf
\EndWhile
\If{ \textcolor{cyan}{\texttt{</retrieve>}}  in $Y$}
    \State Extract retrieval query \( q_r \) from $Y$
    \State Retrieve frame  \( v' = h_r(q_r, \bm{V}) \) 
    \State Insert $v'$ into full response \(Y \gets Y + v' \)
\ElsIf{\textcolor{brown}{\texttt{</generate>}}  in $Y$}
    \State Extract generation query \( q_g \) from $Y$
    \State Retrieve frame  \( v_r' = h_r(q_r, \bm{V}) \) 
    \Comment{As conditional image input for generation}
    \State Generate frame \( v' = h_g(q_g, v_r') \) 
    \State Insert $v'$ into full response \(Y \gets Y + v' \)
\EndIf

\LineComment{\textbf{Round 2: finish reasoning}}
\While{True}
\State Generate response token \( y_t \sim f(\cdot \mid V, Q, A, Y) \)
\State Append \( y_t \) to full sequence \( Y \gets Y + y_t \)
\If{\( y_t \) in [\textcolor{red}{\texttt{</answer>}}, \texttt{<eos>}]}
    break   
\EndIf
\EndWhile

\State \textbf{return} final full response \( Y \)
\end{algorithmic}
\end{algorithm}

\section{Dataset Description}
\label{sec:dataset_description}
In this section, we provide details of how the data is generated, what datasets are used, what retrieval and generation models are called, and how the crucial data filtering calibrated by ground-truth human annotation is performed.

\subsection{Generating data with Interleaved CoT}
To construct the Supervised-Finetuning (SFT) dataset with interleaved text-frame CoTs, where the frames can be retrieved or generated, we begin by selecting datasets that already contain CoTs. This is critical as our interleaved CoTs can then be quantitatively evaluated against existing ones. During early exploration, we experimented with several video-reasoning datasets \citep{han2025videoespresso, feng2025video, wang2025videorft}
that relied on automatically generated CoTs. However, we found that these CoTs were often noisy, inconsistent with the underlying video content, or structurally misaligned with desired reasoning format. As a result, they were suboptimal as supervision for training and quality-control.

We instead base our dataset on evaluation benchmarks that are carefully curated by human annotators and further filtered by quality control processes to create the SFT dataset. We select the following datasets: \textbf{MINERVA} \citep{nagrani2025minerva} is designed for complex video reasoning, featuring $1,515$ hand-crafted multiple-choice questions across 223 long-form videos (avg. 12 minutes) spanning diverse domains and skill sets. A distinguishing characteristic is the inclusion of detailed, human-annotated reasoning traces (avg. 92 words) for each question. These traces explicitly delineate the intermediate steps required for multi-step inference across various reasoning modalities. \textbf{CausalVQA} \citep{foss2025causalvqa} is a physically grounded dataset utilizing egocentric footage from EgoExo4D \citep{grauman2024ego} to probe causal reasoning capabilities in video models. It comprises 793 paired questions ($1,586$ total items) across 779 video segments, categorized into types including counterfactual, hypothetical, and anticipation. The dataset employs this paired-question structure specifically to mitigate reliance on linguistic priors and enforce genuine causal inference. \textbf{Social Genome} \citep{mathur2025social} is a fine-grained, grounded social reasoning dataset, utilizing 272 videos (4.5 hours) of face-to-face interactions. The dataset contains $1,486$ human-annotated reasoning traces that require models to integrate visual, verbal, and vocal cues for accurate inference. Uniquely, this benchmark emphasizes the necessity of incorporating external contextual knowledge for comprehensive social understanding.

\subsection{Interleaved CoT generation} 

We use Gemini 2.5 Pro as the underlying VLM to construct the SFT dataset, chosen for its strong video understanding capabilities \citep{comanici2025gemini}. The reason we do not directly use the human annotation is that we want to create natural retrieval or generation calls inside the CoTs. We follow Algorithm \ref{alg:interleaved} to get the new video-text interleaved CoTs. For retrieval, we use TFVTG \citep{zheng2024training}, a frame retrieval method which relies on the pre-trained large vision model BLIP-2 \citep{li2023blip} and analyzes multiple sub-events and their relationships from the query text. It returns multiple adjacent frames, and we select the middle frame as the retrieved output. The retrieved frame is different from the initial frames as TFVTG searches from the video at a higher sampling rate (3 fps as opposed to 1 fps for the initial input frames). For generation, we first use the same retrieval process to get a frame as the conditional for generation. Then, we use Stable Diffusion 3.5 Large \citep{esser2024scaling} which takes in the text query and the frame tokens separately encoded by the corresponding VAE of Stable Diffusion 3.5 Large, to generate the frame.

\subsection{Data quality control and filtering} \label{sec:data_quality_control}
For all the generated CoTs, we perform the crucial data quality control and filtering caliberated by the ground-truth human-annotated CoTs. First, we filter out all CoTs that lead to wrong answers, and re-generate the CoTs from the corresponding input video-QA pairs until the correct answers are produced. To save computation, we cap the re-generation to two times. To make sure that the generated CoTs are high-quality, we measure the text embedding similarities of the generated interleaved CoTs with the ground-truth CoTs provided by MINERVA, CausalVQA and Social Genome. The ground-truth CoTs are not used when generating the interleaved CoTs, instead they are used to filter out CoT generations deviating significantly from them. This is because feeding ground-truth to Gemini can result in very low call rate of retrieval or generation tools (Sect. \ref{sec:prompt_with_ground_truths}). To perform the filtering, we compare the generated CoTs with ground-truths and only retain them if the BGE M3-Embedding \citep{chen2024m3} similarity is greater than $80\%$, following the practice from \citet{han2025videoespresso} to compare textual CoT semantic similarity. Lastly, we perform a format check to ensure all the CoTs contain necessary thinking and answering tokens, and also perform a manual check of $100$ to validate the CoT quality, retrieval and generation format following, and correct correspondence between the CoT and the answer. All $100$ samples pass the manual checks. 

\subsection{Dataset composition} 
\begin{figure}[t]
	\centering
  \includegraphics[width=0.6\linewidth]{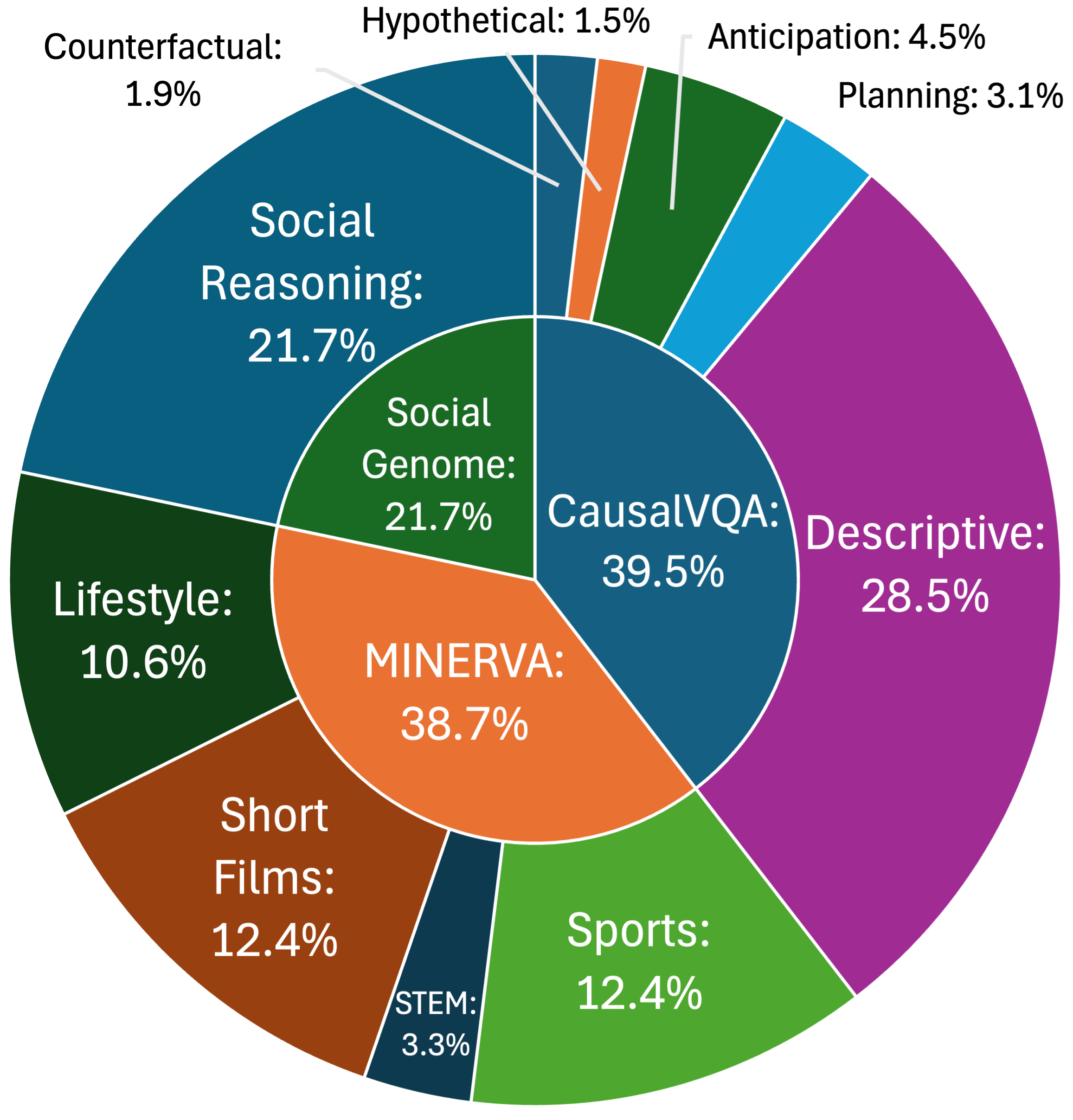}
  \caption{The SFT dataset composition, which uses video from MINERVA \citep{nagrani2025minerva} (multi-step reasoning), CausalVQA \citep{foss2025causalvqa} (physical reasoning), and Social Genome \citep{mathur2025social} (social reasoning).}
  \label{fig:data_pie}
\end{figure}

Based on the steps above, we collect $3,373$ high-quality CoTs, $1,608$ of which contains retrieved or generated frames, taking up $47.67\%$ of the dataset. Out of those, $1,026$ are retrievals and $582$ are generations. As shown in Figure \ref{fig:data_pie}, the dataset contains a diverse set of videos tailored for complex reasoning across tasks, domains and skill sets. Specifically, it contains $1,334 \ (39.5\%)$ video QA pairs from CausalVQA, $1,307\  (38.7\%)$ from MINERVA, and $732\  (21.7\%)$ from Social Genome. Breakdown of sub-categories are in Figure \ref{fig:data_pie}. In Figure \ref{fig:dataset_example}, we show a data instance with text CoT interleaved with a retrieved frame.

\section{Experiments}
Below we show the benchmarks, baselines, training details of the experiments, and show results comparing \Ours with different SOTA video reasoning models, including those sharing similar ideas of adding visual information in the CoT, and multiple ablation studies on how ddifferent SFT data constructions affect performances.
\subsection{Benchmarks} 
We evaluate our approach on five benchmarks designed to probe diverse aspects of video understanding. \textbf{Video-MME}~\citep{fu2025video} assesses general video analysis capabilities across 6 visual domains and 30 fine-grained categories, requiring robustness to highly variable video durations. \textbf{EgoNormia}~\citep{rezaei2025egonormia} introduces the specialized challenge of interpreting physical-social norms and navigating normative decision-making within egocentric contexts, particularly when norms conflict. The next three benchmarks focus specifically on complex reasoning via CoTs. \textbf{VideoEspresso}~\citep{han2025videoespresso} demands fine-grained semantic reasoning across 14 tasks, emphasizing spatial localization and temporal grounding. \textbf{VCR-Bench}~\citep{qi2025vcr} dissects the reasoning process itself, evaluating both perception and logical reasoning capabilities across seven task dimensions. Finally, \textbf{ViTIB}~\citep{zhang2025vitcot} directly assesses the effectiveness and generalizability of the proposed video-text interleaved CoT paradigm—the focus of this work—by requiring the integration of salient key-frames directly into the reasoning.

\begin{table*}[th]
    \centering
    \caption{Performance on video reasoning benchmarks. \Ours consistently outperforms other open-sourced VLM baselines with similar model sizes, and sometimes even outperforms the largest closed-sourced models. The highest is highlighted, and the second highest is underlined for each dataset. $^{\dagger}$ indicates results evaluated by us. Aligning with the original papers, for ViTIB, $\#$ suggests Gemini 2.0 Flash result, $*$ suggests Qwen2.5-VL-7B-Instruct result, and for VideoEspresso, $\triangle$ suggests InternVL2-8B results, respectively. The results are without subtitles for Video-MME, selecting both correct behavior and justification from the verified split for EgoNormia, and average accuracies for VideoEspresso, VCR-Bench, and ViTIB. All results are based on the latest Arxiv versions of the corresponding datasets.}
    \label{tbl:combined_benchmarks_all}
    \resizebox{\textwidth}{!}{
        \begin{tabular}{lcccccccccc}
            \toprule
            \textbf{Model} & \multicolumn{2}{c}{\textbf{Video-MME}} & \multicolumn{2}{c}{\textbf{VideoEspresso}} & \multicolumn{2}{c}{\textbf{EgoNormia}} & \multicolumn{2}{c}{\textbf{VCR-Bench}} & \multicolumn{2}{c}{\textbf{ViTIB}}\\
            \cmidrule(lr){2-3} \cmidrule(lr){4-5} \cmidrule(lr){6-7} \cmidrule(lr){8-9} \cmidrule(lr){10-11}
            & \textbf{Frame} & \textbf{Acc.} & \textbf{Frame} & \textbf{Acc.} & \textbf{Frame} & \textbf{Acc.} & \textbf{Frame} & \textbf{Acc.}  & \textbf{Frame} & \textbf{Acc.}\\
            \midrule
            \multicolumn{11}{c}{\textit{Closed Source Models}} \\
            GPT-4o~\citep{hurst2024gpt} & 1fps & 71.9 & 3fps & 26.4 & 1fps & 45.5 & 64 & 46.9 & - & -\\
            Gemini 2.5 Pro~\citep{comanici2025gemini} & 1fps & \textbf{84.3} & - & - & 1fps & \textbf{64.7} & 1fps & \textbf{61.3} & 1fps & $53.9^{\#}$ \\
            \midrule
            \multicolumn{11}{c}{\textit{Open Source Models}} \\
            Qwen2.5-VL-7B \cite{bai2025qwen2} & 1fps & 65.1 & 1fps & 35.5 & - & - & 64 & 30.4 & 1fps & 49.8*\\
            InternVL2.5-8B \citep{chen2024expanding} & 64 & 64.2 & 1fps & $28.7^{\triangle}$ & 1fps &13.0 & 1fps & 33.0 & 1fps & 56.8\\
            Video-LLaMA3-7B \citep{zhang2025videollama} & 1fps & 66.2 & - & - & - & - & 1fps & 32.5 & 1fps & 51.7\\
            Qwen3-VL-8B-Thinking \citep{bai2025qwen3} & 1fps  & 71.8 & 1fps & $41.5^{\dagger}$ & 1fps & $48.9^{\dagger}$  & 1fps & $38.2^{\dagger}$ & 1fps & $60.2^{\dagger}$ \\
            \rowcolor{lightgray}\textbf{\Ours} & 1fps  & \underline{74.2} & 1fps &  \textbf{46.8} & 1fps  & \underline{51.3} & 1fps & \underline{47.1}  & 1fps  & \textbf{63.3}\\
            \bottomrule
        \end{tabular}
    }
    \label{tab:main_result}
\vspace{-6pt}
\end{table*}

\begin{figure}[t]
	\centering
  \includegraphics[width=1\linewidth]{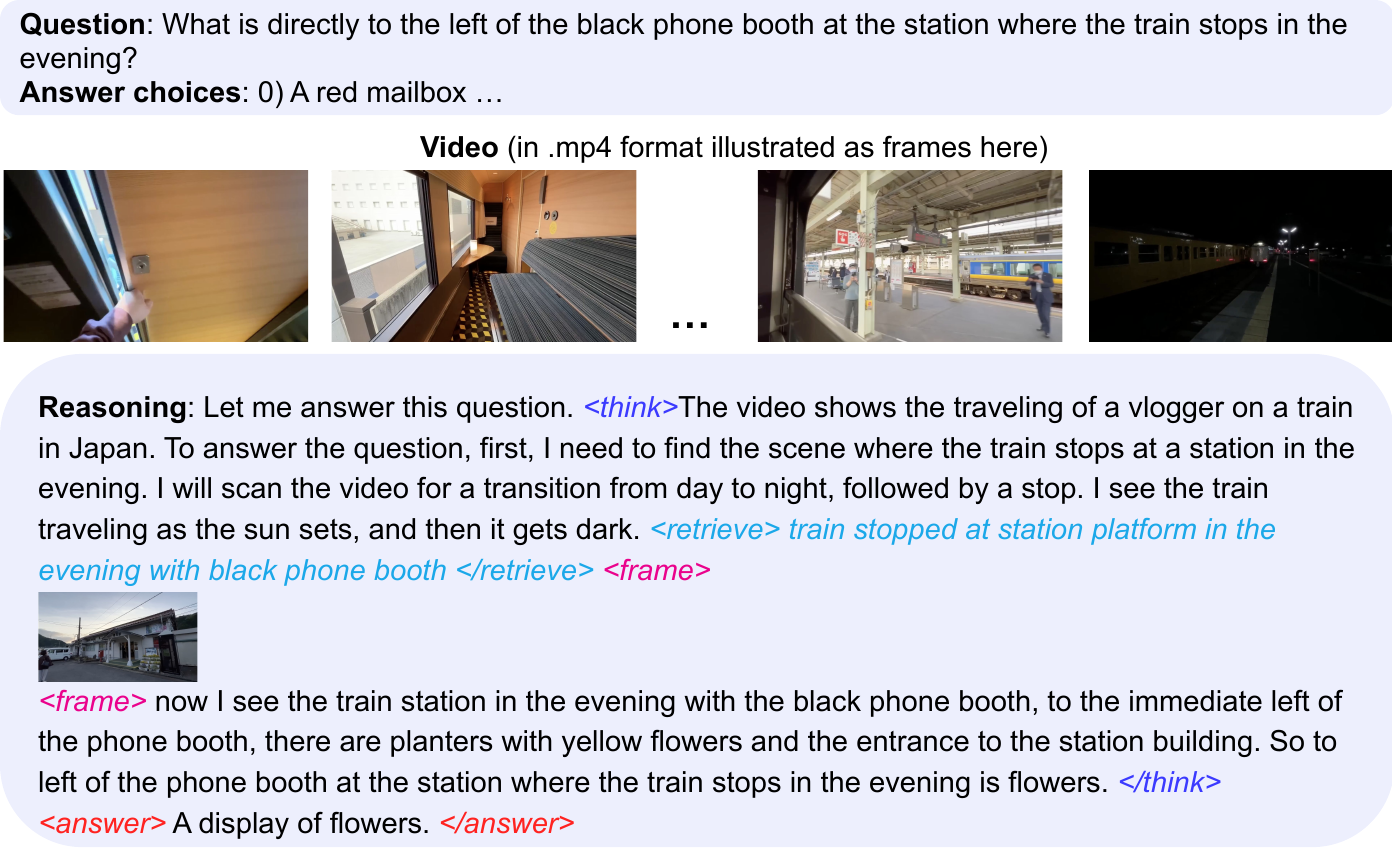}
  \caption{An example of the SFT dataset, consisting of the video, questions and answers, and the frame-text interleaved reasoning.}
  \label{fig:dataset_example}
\end{figure}

\subsection{Baselines}
We compare with different sets of baselines: generic VLMs, video reasoning VLMs with CoTs, and recent video-text interleaving methods. For generic VLMs, we select \textbf{Qwen2.5-VL-7B} \citep{bai2025qwen2}, \textbf{InternVL2.5-8B} \citep{chen2024expanding}, \textbf{Video-LLaMA3-7B} \citep{zhang2025videollama}, and our base model \textbf{Qwen-3-VL-8B-Thinking} \citep{bai2025qwen3}. We also compare with the strongest frontier closd-sourced models, \textbf{GPT-4o} \citep{hurst2024gpt} and \textbf{Gemini 2.5 Pro} \citep{comanici2025gemini}. For video VLMs that are have strong general reasoning capabilities, we include \textbf{Video-R1} \citep{feng2025video} and \textbf{Chain-of-Shot} \citep{hu2025cos}, and \textbf{FrameMind} \citep{ge2025famemind}. For recent video reasoning VLMs with video-text interleaved CoTs, we include \textbf{ViTCoT} \citep{zhang2025vitcot} (inference only), \textbf{Chain-of-Frames} \citep{ghazanfari2025chain} (with SFT), and \textbf{ReWatch-R1} \citep{zhang2025rewatch} (with RL). 

\subsection{Training details}
We leverage Qwen3-VL-8B-Thinking \citep{bai2025qwen3} model for SFT. We fine-tune with LoRA \citep{hu2021lora} with 8 A100 GPUs, using a learning rate of $2.5 \times 10^{-6}$, a batch size of 1, and a single epoch. This setting is also used for all the ablations regarding different SFT data sources. For retrieval and generation at inference time, we use the same TFVTG \citep{zheng2024training} and Stable Diffusion 3.5 Large \citep{esser2024scaling} as the data generation process.

\subsection{Results}

\paragraph{Main results.} From Table \ref{tab:main_result}, \Ours show consistent performance improvements across different video reasoning domains and tasks over other video vision-language models with similar sizes. First of all, \Ours outperforms its base model, Qwen-3-VL-8B-Thinking, overall all five benchmarks, showing that \textit{SFT with interleaved video-text CoT data improves the model's reasoning capabilities on diverse complex video tasks.} 

Second, compared to other similar-sized models (Qwen2.5-VL-7B, InternVL2.5-8B, and Video-LLaMA3-7B), \Ours often show significant improvement, such as VideoEspresso (ours: $46.8$ vs. Qwen2.5-VL-7B: $35.5$, EgoNormia (ours: $51.3$ vs InternVL2.5-8B: $13.0$), and CVR-Bench (ours: $47.1$ vs InternVL2.5-8B: $33.0$), showing that \textit{SFT with interleaved video-text CoTs bring sizable performance gains}.

Lastly, \Ours sometimes compares favorably to the largest closed-source models (GPT-4o and Gemini 2.5 Pro) on video reasoning, and even surpasses GPT-4o on VideoEspresso (ours: $46.8$ vs. GPT-4o: $26.4$) with much fewer frames (ours: 3 vs. GPT-4o: 1), demonstrating the great potential of \textit{leveraging \Ours achieving or even surpassing performances of larger frontier closed-source models.}

\paragraph{Emergent capability of outputting frames in CoT at inference.} We provide a qualitative analysis of some output of \Ours in Figure \ref{fig:inference_examples}. We can see that \Ours enables the model to \textit{dynamically gather new visual information inside the CoT}, either when the key visual frames are not available from the initial frames (factual visual information), or when the key visual information is counterfactual and needs generation (counterfactual). We see that sometimes the generation does not fully reflect the exact prompt (e.g., the last sub-figure in Figure \ref{fig:inference_examples}, where the coffee beans are supposed to be poured into the coffee machine instead of the mug), and we expect \Ours to continue to improve performance based on better generation tools.

\paragraph{Comparison with recent video reasoning CoT methods.} In Table \ref{tab:video_mme}, we compare \Ours with other recent video reasoning baselines focusing on improving reasoning on complex video tasks, some of them also using frames or frame IDs inside CoTs (such as Chain-of-Shot, FrameMind and Chain-of-Frames). \Ours outperforms other video reasoning baselines, except the concurrent work Chain-of-Frames, showing the effectiveness of our approach.
\begin{table}[th]
\caption{Accuracies on Video-MME compared with methods focusing on improving video reasoning, especially those interleaving text and video, such as FrameMind and Chain-of-Frames. \Ours outperforms other video reasoning methods except concurrent work Chain-of-Frames, demonstrating the effectiveness of leveraging frames in CoT.}
\begin{adjustbox}{max width=0.45\textwidth}

\begin{tabular}{ccccc}
\toprule
\textbf{Model} & \textbf{Base Model} & \textbf{Frame}  & \textbf{Acc.} \\
\midrule
Video-R1~\citep{feng2025video} & Qwen2.5-VL-7B-Instruct & 64 & 61.4  \\
Chain-of-Shot \citep{hu2025cos} & LLaVA-Video & 64  & 64.4 \\
FrameMind \citep{ge2025famemind} & Qwen2.5-VL-7B & 32  & 60.9 \\
Chain-of-Frames \citep{ghazanfari2025chain} & InternVL3-8B & 1fps & \textbf{75.3} \\
Qwen3-VL-8B-Think \citep{bai2025qwen3} & Qwen3-VL-8B-Think & 1fps & 71.8 \\
\rowcolor{lightgray} \Ours & Qwen3-VL-8B-Think & 1fps & \underline{74.2} \\
\bottomrule
\end{tabular}
\end{adjustbox}
\label{tab:video_mme}
\vspace{-10pt}
\end{table}
\paragraph{SFT vs. inference only video-text interleaving.} We compare \Ours with an inference-only video-text interleaving CoT method, ViTCoT \citep{zhang2025vitcot}, which prompts a model for initial reasoning, inserts a pre-selected keyframe, and uses the initial reasoning and keyframe as pretext for final reasoning. Different from ViTCoT, we have the SFT phase and dynamically retrieve or generate frames instead of using pre-selected key frames. From Table \ref{tab:sft_vs_inference}, we observe clear improvements of SFT over inference-only interleaving on ViTIB and Video-MME, demonstrating that \textit{SFT focusing on dynamic active perception can increase the video reasoning capability more than inference-only interleaving}.
\begin{table}[th]
\caption{Performance comparison of inference-only ViTCoT and SFT-based \Ours. $^{\dagger}$ indicates results evaluated by us. SFT shows consistent improvement over inference-only video-text interleaving CoT on video reasoning tasks.}
\begin{adjustbox}{max width=0.45\textwidth}

\begin{tabular}{ccccc}
\toprule
\textbf{Model} & \textbf{Base Model} & \textbf{Frame}  & \textbf{ViTIB} & \textbf{Video-MME} \\
\midrule
ViTCoT \citep{zhang2025vitcot} & InternVL2.5-8B & 1fps & 58.2 & 66.8$^{\dagger}$ \\

ViTCoT \citep{zhang2025vitcot} & Qwen3-VL-8B-Think & 1fps & 59.9$^{\dagger}$ & 70.2$^{\dagger}$ \\
\rowcolor{lightgray} \Ours & Qwen3-VL-8B-Think & 1fps & \textbf{63.3} & \textbf{74.2} \\
\bottomrule
\end{tabular}
\end{adjustbox}
\label{tab:sft_vs_inference}
\vspace{-10pt}
\end{table}

\begin{figure*}[th]
	\centering
  \includegraphics[width=\linewidth]{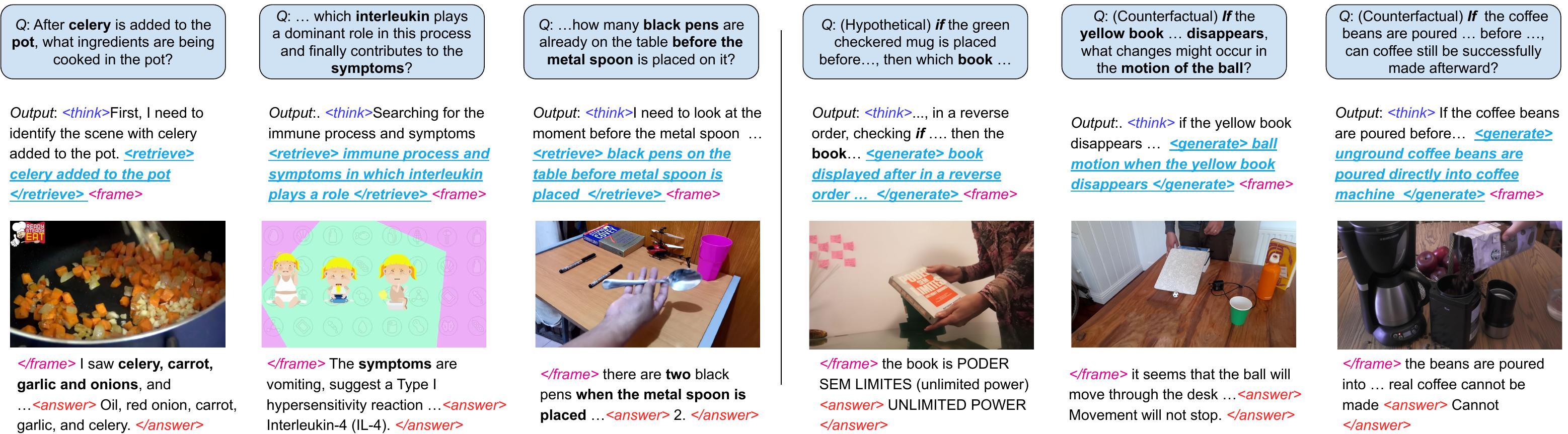}
  \caption{Qualitative examples showing the emergent capability of \Ours outputting retrieved (left half) or generated frames (right) inside the CoTs at inference time, dynamically adding requested visual information (factual original videos via retrieval or counterfactual synthesis via conditional image generation) to the reasoning process.} 
  \label{fig:inference_examples}
\vspace{-8pt}
\end{figure*}

\paragraph{SFT vs. RL-based video-text interleaving.} We also compare \Ours with recent RL-based video-text interleaving methods, ReWatch-R1 \citep{zhang2025rewatch} and FrameMind \citep{ge2025famemind}, on VCR-Bench and Video-MME, respectively. From Table \ref{tab:sft_vs_rl}, we also see consistent improvement over RL-based method with the exact same frame sampling rate and base models. We suspect this is a data quality issue, as ReWatch-R1 \citep{zhang2025rewatch} pointed out lower-quality data can degrade performance of RL or CoT significantly. We conduct further ablation of data quality control in paragraph \ref{para:data_quality}, which confirms that poorer quality CoT data (often generated instead of human-annotated) hurt performance significantly.
\begin{table}[th]
\caption{Performance comparison of SFT-based \Ours and RL-based video-text interleaving CoT methods, Rewatch-R1 and FrameMind. For fair comparison, we also conducted SFT on Qwen2.5-VL-7B and evaluated on VCR-Bench and VideoMME. The result of \Ours of Video-MME differ from Table \ref{tab:main_result} because of frame sampling change to match FrameMind. \Ours shows consistent improvements over RL-based methods on complex video reasoning tasks.}
\begin{adjustbox}{max width=0.45\textwidth}

\begin{tabular}{cccc}
\toprule
\textbf{Model} & \textbf{Base Model} & \textbf{Frame}  & \textbf{Acc.}\\
\midrule
\multicolumn{4}{c}{\textit{VCR-Bench \citep{qi2025vcr}}} \\
ReWatch-R1 \citep{zhang2025rewatch} & Qwen2.5-VL-7B & 1fps & 39.6 \\
\Ours & Qwen2.5-VL-7B & 1fps & 42.2 \\
\rowcolor{lightgray} \Ours & Qwen3-VL-8B-Think & 1fps & \textbf{47.1} \\
\midrule
\multicolumn{4}{c}{\textit{Video-MME \citep{fu2025video}}} \\

FrameMind \citep{ge2025famemind} & Qwen2.5-VL-7B & 32 & 60.9 \\
\Ours & Qwen2.5-VL-7B & 32 & 65.8 \\

\rowcolor{lightgray} \Ours & Qwen3-VL-8B-Think & 32 & \textbf{72.4} \\
\bottomrule
\end{tabular}
\end{adjustbox}
\label{tab:sft_vs_rl}
\vspace{-6pt}
\end{table}

\subsection{Ablation studies}
Below we include ablations regarding CoT sources and tool uses, and more analyses can be found in Appendix. 

\paragraph{Prompting with or without ground-truth CoT.}  We created two small SFT datasets, each of $1,000$ samples, to see if including ground-truth CoT when prompting Gemini to create the CoT for SFT can help or hurt performance. From Table \ref{tab:prompt_with_ground_truths}, prompting with ground-truth will significantly lower the call rate of retrieval or generation tools, and we also observe the downstream performance degrade as a result. Following this experiment, we do not feed the ground-truth into the original prompt to increase the chance of Gemini calling the retrieval or generation tool calls, with the trade-off of the generated CoT text may not exactly follow the ground-truth. \textit{This suggests that inclusion of visual information can make up for the slight deviation of generated CoT texts from ground-truth CoTs.}
\begin{table}[th]
\caption{Performance comparison of small $1,000$-sample SFT with including vs. not including ground-truth to prompt Gemini 2.5 Pro. The results suggest that including CoT significantly decreases the chance of Gemini 2.5 Pro calling retrieval or generation tools, and the lower frame inclusion rate leads to lower downstream performance.}
\begin{adjustbox}{max width=0.45\textwidth}

\begin{tabular}{cccc}
\toprule
\textbf{CoT generation} & \textbf{Ground Truth CoT} & \textbf{$\%$ CoT w/ frame}  & \textbf{Video-MME} \\
\midrule

Retrieval+Generation & Included & $5.26\%$ & 72.2 \\
\rowcolor{lightgray} Retrieval+Generation & Not Included & $45.21\%$ & \textbf{73.7} \\

\bottomrule
\end{tabular}
\end{adjustbox}
\label{tab:prompt_with_ground_truths}
\vspace{-14pt}
\end{table} \label{sec:prompt_with_ground_truths}

\paragraph{Ablation of different CoT data sources.} \label{para:data_quality} We compared two sets of datasets with ground-truth CoTs as the SFT sources, where we created a $1,000$ sample SFT dataset based on CoTs 1) generated by VLMs (from VideoEspresso \citep{han2025videoespresso} and Video-R1 \citep{feng2025video} training sets), or 2) CoTs annotated by humans (our approach; from MINERVA \citep{nagrani2025minerva}, CausalVQA \citep{foss2025causalvqa}, and Social Genome \citep{mathur2025social}). Note that we still do not directly use the ground-truth CoT for prompting; instead, we compare the generated CoTs with ground-truths and retain them if the BGE M3-Embedding similarity is greater than $80\%$, similar to what we have done in Sect. \ref{sec:data_quality_control}. From Table \ref{tab:cot_data_source_ablation}, we see a significant performance difference between the VLM-generated data source and human-annotated data source, and notably the results after SFT from VLM-generated data sources can be even worse than the base model (row Qwen3-VL-8B-Think in Table \ref{tab:main_result}. \textit{This shows the importance of SFT source data CoT quality for improving model performance.}
\begin{table}[th]
\caption{Performance comparison of small $1,000$-sample SFT by prompting from datasets with VLM-generated or human-annotated ground-truth CoTs. Results are different from \ref{tab:main_result} because of only $1,000$ samples. FPS is 1. There is a performance jump from VLM-generated to human-annotated, suggesting that source data quality is important.}
\begin{adjustbox}{max width=0.45\textwidth}

\begin{tabular}{ccccc}
\toprule
\textbf{CoT source} & \textbf{Data} & \textbf{Base Model}  & \textbf{ViTIB} & \textbf{Video-MME}\\
\midrule
VLM-generated & \begin{tabular}{@{}c@{}}VideoEspresso \\ Video-R1\end{tabular} & Qwen3-VL-8B-Think & 58.3 & 68.6 \\
\rowcolor{lightgray} Human-annotated & \begin{tabular}{@{}c@{}}MINERVA \\ CausalVQA \\ Social Genome\end{tabular} & Qwen3-VL-8B-Think & \textbf{62.0} & \textbf{73.7} \\

\bottomrule
\end{tabular}
\end{adjustbox}
\label{tab:cot_data_source_ablation}
\vspace{-12pt}
\end{table}

\paragraph{Ablations of text-only, retrieval-only, generation-only, and retrieval and generation SFT data.} We created four small SFT datasets, each of $1,000$ samples, to see if interleaving with retrieved and generated frames indeed help. The four sets are: text-only CoT SFT,  retrieval-only CoT SFT, generation-only CoT SFT, and text interleaved with retrieval or generation frames (our setting). We use Gemini 2.5 Pro to generate all four small datasets. We use the same SFT recipe for all cases. From Table \ref{tab:retrieval_or_generation_only}, either retrieval or generation helps improve the performance on Video-MME over text-only CoT, but blending both as \Ours performed the best, suggesting the benefit of using both visual retrieval and generation calls in creating the SFT dataset.
\begin{table}[th]
\caption{Performance comparison of small $1,000$-sample SFT with text-only CoT, retrieval-only CoT, generation-only CoT, and interleaved retrieval/generation frames CoT data (our setting). The results suggest that using retrieval and generation achieves better results than using either tool only.}
\begin{adjustbox}{max width=0.45\textwidth}

\begin{tabular}{cccc}
\toprule
\textbf{CoT generation} & \textbf{Base Model} & \textbf{Frame}  & \textbf{Video-MME} \\
\midrule
Text-only & Qwen3-VL-8B-Think & 1fps & 71.9 \\
Retrieval-only & Qwen3-VL-8B-Think & 1fps & 72.8 \\
Generation-only & Qwen3-VL-8B-Think & 1fps & 72.4 \\
\rowcolor{lightgray} All (ours) & Qwen3-VL-8B-Think & 1fps & \textbf{73.7} \\

\bottomrule
\end{tabular}
\end{adjustbox}
\label{tab:retrieval_or_generation_only}
\vspace{-10pt}
\end{table}


\section{Related Work}
\paragraph{Video reasoning models.} As VLMs are increasingly applied to video understanding and reasoning \citep{lee2025interactive,liang2024foundations}, recent work has been adapting CoT methods specifically for video understanding, either by adding explicit textual reasoning explanations or by identifying relevant visual features. Video-R1 \citep{feng2025video} uses synthesized SFT and RL data to improve video reasoning. VideoCoT \citep{wang2024videocot} introduced a benchmark and annotation tool to generate reasoning explanations. Video-of-Thought \citep{fei2024video} introduces a complex pipeline that decomposes tasks into subproblems moving from pixel-level perception to high-level reasoning via spatial-temporal scene graphs. Video-RFT \citep{wang2025videorft} proposes higher-quality SFT and RL datasets by using rich, structured literal representations and condition the generation on the actual video. Chain-of-Shot \citep{hu2025cos} introduces an adaptive shot selection method framed as test-time optimization.

\paragraph{Video frame information in CoTs and active visual perceptions.} To further improve video CoTs, recent work starts to leverage video frame information inside the CoT. ViTCoT \citep{zhang2025vitcot} is an inference-only methods that first extracts keyframes from videos offline, then deploys a two-round inference, where the first round prompts the model to call retrieval, and the second round appends the previous response with the retrieved keyframe for final answer. Chain-of-Frames \citep{ghazanfari2025chain} designs CoTs for SFT which explicitly reference relevant frame ID numbers. FrameMind \citep{ge2025famemind} employs RL to enable video models to dynamically zoom in frames or scan an interval of multiple frames in the CoT. ReWatch-R1 \citep{zhang2025rewatch} uses SFT and RL, based on synthetic video-grounded Chain-of-Thought data by simulating a re-watching process and adding frame time stamps. Works such as FrameMind, ReWatch-R1 and others \citep{su2025openthinkimg, zheng2025deepeyes} share the active perception perspective \citep{bajcsy1988active}, where the models actively gather new visual information based during reasoning or prediction. Improving upon these, our work enables the emergent capability of active perception with both frame generation and retrieval for video reasoning tasks. 
\vspace{-6pt}
\section{Conclusion}
\vspace{-2pt}
We introduced Act to See (\Ours), a novel framework that enhances Vision-Language Models (VLMs) by enabling active visual perception for complex video reasoning. By integrating specialized tool calls within the CoT process, and curating a high-quality SFT dataset filtered against human annotations, \Ours allows models to dynamically retrieve existing frames or generate hypothetical scenarios in CoTs, overcoming the limitations of static inputs and suboptimal reasoning. Extensive evaluations across five diverse video reasoning benchmarks demonstrate the effectiveness of \Ours, showing significant improvements over SOTA VLMs of comparable-sized video CoT methods. We observe the emergent ability of the model to actively search for or synthesize visual information inside reasoning traces, marking an advancement toward complex and causally grounded video reasoning.
\paragraph{Acknowledgement} The research is supported by National Institutes of Health awards R01MH125740, R01MH132225, R21MH130767, ONR N000142312368, ONR MURI N00014-25-1-2116, as well as the Amazon AI PhD Fellowship.

{
    \small
    \bibliographystyle{ieeenat_fullname}
    \bibliography{main}
}



\clearpage
\setcounter{page}{1}
\maketitlesupplementary





\def\paperID{} 
\def\confName{CVPR}
\def\confYear{2026}



\appendix


\section{Impact statement}

The \Ours framework significantly advances complex video reasoning by empowering models with active visual perception, enabling them to dynamically retrieve or synthesize visual information. However, we acknowledge the potential negative societal impacts associated with these capabilities. A significant concern arises from the framework’s ability to generate novel visual frames to illustrate hypothetical or counterfactual scenarios. This generative function could be exploited by malicious actors to synthesize convincing deceptive media (deepfakes) or fabricate visual ``evidence'', thereby facilitating fraud, harassment, or the manipulation of public discourse. Furthermore, the enhanced efficiency and granularity of video analysis, particularly the active retrieval mechanism, could be misused to develop or extend harmful forms of surveillance. This risks intensifying the monitoring and profiling of individuals without consent, raising substantial privacy and human rights concerns. Additionally, the framework may inherit and perpetuate societal biases present in the underlying foundational models used for data generation and frame synthesis. To mitigate these risks, we strongly advocate for the development of robust forensic tools and digital watermarking to ensure the traceability of synthesized content. We also recommend implementing strict ethical guidelines for deployment, including rigorous bias evaluations and potentially gated access to the generative components, to prevent misuse in sensitive domains.

\section{SFT synthesis prompt}
Below we include the full prompts for round 1 (Table \ref{tab:first_round_instruction}) and round 2 (Table \ref{tab:second_round_instruction}) each with an example included. Round 1 prompts Gemini \citep{comanici2025gemini} to output reasoning traces with retrieval or generation requests along with the corresponding queries. Then the query will be used to retrieve or generate with offline retrieval and generation models. The text after \textcolor{cyan}{\texttt{</retrieve>}} or \textcolor{brown}{\texttt{</generate>}} will be removed, the retrieved or generated frames will be added, and then altogether these will be input for the second round. The second round is to let Gemini finish the reasoning and output the answer based on the extra visual information. No ground truth answer or reasoning is included to encourage diversity of the response.
\begin{table*}[h]
    \centering
    \begin{tabular}{p{13cm}}
        \hline
        Provide a step-by-step answer for the question below. You will be provided a set of initial video frames. 
        Your task is to generate the reasoning traces, which include texts and images, to correctly answer the question.
        You must conduct reasoning inside \think{and}. When reasoning, if you need any visual
        knowledge, you can call a frame retrieval module by \retrieve{\textit{retrieval prompt}},
        or a frame generation module by \generate{\textit{generation prompt}}, and either will
        return the video frame between \frames{and}. The retrieval or generation prompt is the text description of the frame you want to retrieve/generate and you should provide it with specificity and conciseness about the exact content of the image, instead of timestamps (e.g., 2 seconds) in the query. Please do not include any timestamps in the query. Example query: ``\textit{a bison baby chased by a wolf pack}''.
        Try calling the module whenever needed. Finally, give the answer in free-form text in the end, with \answer{and} tags. \\
        Please following the format below: \\
        {\textcolor{blue}{\texttt{<think>}}} \\
            reasoning steps \\
            \retrieve{\textit{retrieval prompt}} (or \generate{\textit{generation prompt}}) \\
            reasoning step \\
        {\textcolor{blue}{\texttt{</think>}}}\\
        \answer{\textit{Answer to the question in free-form text.}}\\

        Question: ``What is directly to the left of the black phone booth at the station where the train stops in the evening?'' \\
        Answers:  ``0'': ``A red mailbox.'', \\
        ``1'': ``A couple of steps.'', \\
        ``2'': ``A few white vans.'', \\
        ``3'': ``A telephone pole.'', \\
        ``4'': ``A display of flowers.'',\\
        Video: [video file]\\
        
        \hline
    \end{tabular}
    \caption{An example of Gemini input to collect SFT data for \Ours in the first round. This is to get the start of the reasoning along with the retrieval or generation queries. Anything after the retrieval prompt or generation prompt will be discarded.}\label{tab:first_round_instruction}
\end{table*}

\begin{table*}[h]
    \centering
    \begin{tabular}{p{13cm}}
        \hline
        Provide a step by step answer for the question below. You will be given a half-finished reasoning steps
        you provided last round, which asked for new additional frames. You will then be provided the set of initial frames and the additional frames you asked for. 
        Your task is to generate the rest of the reasoning traces in text correctly answer the question. Do NOT include any \retrieve{...} or \generate{...} this time, as these modules are not available now.
        You must finish the reasoning started from the given half-finished response and finish the reasoning with {\textcolor{blue}{\texttt{</think>}}}. Start the reasoning directly without {\textcolor{blue}{\texttt{<think>}}}. 
        Finally, give the answer in free-form text in the end, with \answer{and} tags.
        Please following the format below: \\
            reasoning step \\
        {\textcolor{blue}{\texttt{</think>}}}\\
        \answer{\textit{Answer to the question in free-form text.}} \\
        Question: ``What is directly to the left of the black phone booth at the station where the train stops in the evening?'' \\
        Answers:  ``0'': ``A red mailbox.'', \\
        ``1'': ``A couple of steps.'', \\
        ``2'': ``A few white vans.'', \\
        ``3'': ``A telephone pole.'', \\
        ``4'': ``A display of flowers.'',\\
        Half-completed reasoning: \\
        Let me answer this question. {\textcolor{blue}{\texttt{<think>}}} The question asks what is directly to the left of the black phone booth where the train stops in the evening. The video is a vlog based on a trip of the train, where the train arrives when it is dark. To know the answer, I need to retrieve the visual details when the train stops in the evening. \retrieve{train stopped at station platform in the evening with black phone booth}\\
        \frames{[retrieved video frame]}\\

        \hline

            \end{tabular}
    \caption{An example of Gemini input to collect SFT data for \Ours in the second round. This is to complete the reasoning from the first round.}\label{tab:second_round_instruction}
\end{table*}


\section{Dataset}
Below we include more details of the three datasets we used to create the SFT data:

\paragraph{MINERVA} (Multimodal INterpretablE Reasoning Video Annotations) \citep{nagrani2025minerva} is a challenging benchmark designed to rigorously evaluate the intermediate reasoning capabilities of multimodal models. Traditional video benchmarks often rely solely on outcome supervision, obscuring whether models genuinely integrate perceptual and temporal information or exploit biases. MINERVA addresses this by providing 1,515 complex, hand-crafted, multi-step questions across diverse video domains and lengths (up to 100 minutes), each accompanied by detailed, manually annotated reasoning traces. These traces facilitate interpretable assessment beyond final answer accuracy by outlining the necessary localization, perception, and logical steps required for the solution. Extensive evaluation demonstrates that MINERVA poses a significant challenge to frontier models (best performance $66.2\%$ vs. $92.5\%$ human accuracy), and an accompanying taxonomy of errors reveals that temporal localization and visual perception remain primary failure modes. Type of the videos include Short Films, Sports and Board Games, Educational, and Lifestyle. Type of skills required for the QA include Temporal Reasoning, Counting, Cause and
Effect, Goal Reasoning, Situational Awareness, Event Occurrence,
State Changes, Reading (OCR), Listening (identifying
a detail in the audio track), Spatial Perception, Numerical
Reasoning (all math operations other than counting), Object
Recognition, Counterfactual Reasoning (‘what if’, but with
an objective outcome).

\paragraph{Social Genome} \citep{mathur2025social} is the first benchmark designed to evaluate the fine-grained, grounded social reasoning capabilities of multimodal models. The dataset comprises 272 videos of real-world social interactions, accompanied by 1,486 human-annotated reasoning traces totaling 5,777 steps. Each reasoning step is explicitly grounded in multimodal evidence, referencing visual, verbal, and vocal cues extracted from the interactions. Uniquely, SOCIAL GENOME is the first benchmark to systematically incorporate external knowledge—contextual information not present in the video stimuli—which accounts for $51\%$ of the reasoning steps. This dense annotation structure, featuring over $11,000$ entities and $2,900$ external knowledge observations, facilitates a holistic evaluation of the semantic and structural validity of model-generated social reasoning traces.

\paragraph{CausalVQA} \citep{foss2025causalvqa} is a benchmark dataset designed to evaluate multimodal models' capacity for physically grounded causal reasoning in real-world scenarios. Existing Video Question Answering (VQA) benchmarks typically focus on surface-level perception or rely on narrow synthetic simulations; CausalVQA bridges this gap by utilizing egocentric videos sourced from the EgoExo4D dataset. The benchmark comprises $1,586$ items (793 paired questions) across five categories designed to probe causal understanding: counterfactual, hypothetical, anticipation, planning, and descriptive. To ensure robustness, the dataset was curated using a rigorous hybrid human-and-model pipeline specifically engineered to enforce visual grounding and mitigate reliance on linguistic shortcuts, incorporating mechanisms such as immunization against "blind" LLMs and the use of paired, perturbed distractor sets. Baseline evaluations reveal a substantial gap between state-of-the-art models ($61.66\%$) and human performance ($84.78\%$), particularly on anticipation and hypothetical questions, underscoring the challenge of applying spatial-temporal reasoning and physical principles in complex, real-world settings. The types of questions in CausalVQA include Counterfactual, Hypothetical, Anticipation, Planning, and Descriptive.

\subsection{Details of the retrieval and generation tool}
Below we provide the brief introductions to the retrieval and generation tools we use. In \Ours, we simply use the original video as the input, and use the retrieval model to retrieve a series of relevant frames based on the retrieval or generation query provided by Gemini. For retrieval requests, we use the middle frame of retrieved frames. For generation requests, we use the middle frame of retrieved frames as conditional, and use the generation query to generate the frame.

\paragraph{TFVTG} (Training-Free Video Temporal Grounding) is an approach that synergizes the reasoning capabilities of Large Language Models (LLMs) with the alignment strengths of Vision-Language Models (VLMs), eliminating the reliance on annotated training data. To handle complex, multi-event queries, TFVTG first employs an LLM (BLIP-2 \citep{li2023blip}) to decompose the query into constituent sub-events and infer their temporal order and relationships. Crucially, to overcome the tendency of VLMs to overlook dynamic transitions, TFVTG propose a novel localization mechanism that explicitly models both the dynamic transition and static status phases of an event. This mechanism utilizes distinct dynamic and static scoring functions to measure the rate of similarity change and comparative relevance, respectively. Finally, the localized proposals for each sub-event are filtered and integrated based on the LLM-derived temporal constraints. This framework ensures more comprehensive event boundary localization and demonstrates superior generalization capabilities across diverse datasets and out-of-distribution scenarios.

\paragraph{Stable Diffusion 3.5} is an open family of latent diffusion models that introduce architectural and training refinements to improve prompt adherence, image quality, and controllability while remaining efficient on consumer-class hardware. 
The models incorporate Query-Key Normalization within transformer blocks and an enhanced MMDiT-X backbone, which together stabilize training and make the base models more amenable to downstream fine-tuning and multi-resolution generation. 
 The model emphasizes customizability and diversity of outputs, deliberately allowing higher intra-prompt variability to preserve a broader style and knowledge distribution in the base models. 

 Distilled and medium-capacity variants extend the design to latency- and resource-constrained settings, targeting competitive text–image alignment and visual fidelity within a more accessible computational and licensing regime. 
\subsection{Data quality control and filtering}
For all the generated CoTs, we perform quality check below. In the first step, we filter out all CoTs that lead to wrong answers, and re-generate the CoTs from the corresponding input video-QA pairs until the correct answers are produced. We cap the re-generation to two times. To make sure that the generated CoTs are high-quality, we measure the text embedding similarities of the generated interleaved CoTs with the ground-truth CoTs provided by MINERVA, CausalVQA and Social Genome. Specifically, we compare the generated CoTs with ground-truths and only retain them if the BGE M3-Embedding \citep{chen2024m3} similarity is greater than $80\%$, following the practice from \citet{han2025videoespresso}. Lastly, we perform a format check to ensure all the CoTs contain necessary thinking and answering tokens, and remove CoTs without the thinking tokens or answer tokens. Lastly, we perform a manual check of $100$ to validate the CoT quality, where each CoT is inspected by checking whether the retrieval and generation formats are followed, if the reasoning in the CoT and the answer are consistent and coherent, and if the automatic format check and answer check is correct. All $100$ samples pass the manual checks. 

\subsection{Metadata}
After quality control and filtering, we have $3,373$ samples in the SFT dataset. There are $1,334$ from CausalVQA, $1,307$ from MINERVA, and $732$ from Social Genome. To be specific, there are $64$ samples of Counterfactual, $50$ of Hypothetical, $152$ of Anticipation, $106$ of Planning, and $962$ of Descriptive from CausalVQA, and $418$ of Sports, $112$ of STEM, $418$ of Short Films, $359$ of Lifesyle from MINERVA.
\section{Experimental details}
Below we include additional details of the experiments including benchmarks and baselines.
\subsection{Benchmark details}
Below we provide brief introductions to the benchmarking datasets we used in this paper.

\paragraph{Video-MME} \citep{fu2025video} is a comprehensive video benchmark designed to assess MLLMs in video analysis. It comprises 900 manually curated videos totaling 254 hours and $2,700$ multiple-choice questions. The dataset contains diverse content spanning 6 visual domains and 30 fine-grained categories, with video durations ranging widely from 11 seconds to 1 hour.

\paragraph{VideoEspresso} \citep{han2025videoespresso} is a large-scale dataset focused on fine-grained video reasoning via Chain-of-Thought (CoT). Constructed using an automatic pipeline, it features Video-QA pairs enriched with multimodal CoT annotations, including intermediate reasoning steps, spatial bounding boxes, and temporal grounding. The benchmark evaluates models across 14 diverse tasks, emphasizing complex semantic reasoning while preserving spatial details and temporal coherence.

\paragraph{Egonormia} \citep{rezaei2025egonormia} is a benchmark designed to evaluate the understanding of physical-social norms (PSNs) in Vision-Language Models. It comprises $1,853$ multiple-choice questions grounded in $1,077$ unique egocentric videos sourced from Ego4D. The benchmark spans seven norm categories (e.g., safety, privacy, proxemics, politeness, cooperation, coordination/proactivity, and communication/legibility) and focuses on evaluating normative decision-making, particularly in situations where norms conflict.

\paragraph{VCR-Bench} \citep{qi2025vcr} is a comprehensive evaluation framework dedicated to Video Chain-of-Thought (CoT) Reasoning. It consists of 859 videos and 1,034 question-answer pairs (including both multiple-choice and open-ended formats), featuring 4,078 manually annotated reference reasoning steps. The benchmark assesses the entire reasoning process across seven distinct task dimensions, distinguishing between models' perception and logical reasoning capabilities.

\paragraph{ViTIB} \citep{zhang2025vitcot}, the Video-Text Interleaved Benchmark, is constructed to support the Video-Text Interleaved CoT (ViTCoT) paradigm. Sourced from VideoEspresso, it contains $1,382$ videos across 14 categories. The benchmark features $5,051$ MLLM-selected and manually verified key-frames (averaging 3.7 per key-video), designed to be integrated directly into the reasoning process to facilitate intuitive video comprehension. This is the benchmark that directly assess the effectiveness and generalizability of the video-text interleaved CoT paradigm, the key focus of this paper.

\subsection{Baseline details}
Below we provide brief introductions of the baselines we used in this paper.

\paragraph{Qwen2.5-VL-7B} \citep{bai2025qwen2} integrates a dynamic-resolution Vision Transformer with the Qwen2.5 language model via an MLP-based merger to facilitate efficient token compression. The vision encoder utilizes 2D/3D patching, window attention, and SwiGLU to process inputs at near-native resolution while utilizing absolute image coordinates for precise spatial grounding. Temporal reasoning is achieved through Multimodal Rotary Position Embeddings (M-RoPE) aligned to absolute time, enabling robust handling of variable resolutions and video lengths. Finally, a staged training regimen involving large-scale multimodal pre-training and preference optimization yields advanced capabilities in document parsing, object grounding, and agentic interaction.
\paragraph{Intern 2.5-8B} \citep{bai2025qwen2} implements an optimized scaling methodology within a ViT-MLP-LLM framework, utilizing a Progressive Scaling Strategy to align large-scale vision encoders with language models via incremental learning. To efficiently manage diverse inputs such as multi-image and video data, the architecture incorporates dynamic high-resolution processing and multimodal data packing. Furthermore, a rigorous data filtering pipeline is employed to mitigate noise and repetition, thereby enhancing training stability and the efficacy of test-time scaling techniques like CoT reasoning.

\paragraph{Video-LLaMA3-7B} \citep{zhang2025videollama} establishes a vision-centric multimodal foundation by extending robust image understanding to the video domain through a four-stage training pipeline that leverages large-scale, re-captioned image–text corpora and curated temporal data. The architecture utilizes Any-resolution Vision Tokenization with a ViT-based encoder and 2D RoPE to process inputs at arbitrary resolutions and aspect ratios, ensuring the preservation of fine-grained spatial details. To optimize temporal efficiency, a Differential Frame Pruner compresses video content by eliminating redundant patches between consecutive frames, yielding compact and informative tokens. This unified framework effectively handles diverse inputs, including documents, charts, and both short and long videos, within a single instruction-tuned model.

\paragraph{Qwen3-VL-8B} \citep{bai2025qwen3} integrates a Vision Transformer-based visual encoder with a Qwen3 text backbone, forming a unified autoregressive transformer designed to process interleaved text, images, and videos. The architecture employs Interleaved-MRoPE for spatio-temporal positional encoding, DeepStack fusion for multi-level feature integration, and text–timestamp alignment to facilitate precise temporal grounding and long-context reasoning. Following large-scale multimodal pretraining, instruction tuning yields both dense and Mixture-of-Experts (MoE) variants, offering specialized modes for instruction following and complex reasoning. This design ensures robust performance across diverse tasks, including multilingual OCR, complex visual question answering, and agentic GUI control, while supporting scalable deployment from edge to cloud environments.

\paragraph{Video-R1} \citep{feng2025video} adapts rule-based reinforcement learning to video reasoning tasks by combining a temporal-aware optimization algorithm with mixed image–video training data. Central to this approach is Temporal Group Relative Policy Optimization (T-GRPO), which isolates and rewards explicit temporal reasoning by contrasting model performance on ordered versus shuffled video frames. The training protocol proceeds through supervised fine-tuning on a Chain-of-Thought dataset followed by reinforcement learning on a verifiable-answer dataset, employing a length-based reward to balance reasoning depth and succinctness. This methodology results in a model exhibiting enhanced temporal understanding and robust generalization across various video understanding benchmarks.

\paragraph{Chain-of-Shot} \citep{hu2025cos} serves as a training-free, test-time mechanism for video MLLMs that enhances long-video comprehension through adaptive visual input optimization relative to a specific query. The method employs a binary video summarization technique wherein mosaiced shots are evaluated by a general MLLM to generate binary codes that pseudo-temporally ground task-relevant segments. Subsequently, balanced sequences of positive and negative sub-shots are processed alongside original shots via a co-reasoning module that dynamically reweights token logits based on the sparsity of relevant content. This design effectively concentrates computation on informative features while suppressing noise, enabling robust reasoning across varying video lengths without requiring architectural modifications.

 \begin{figure*}[th]
	\centering
  \includegraphics[width=\linewidth]{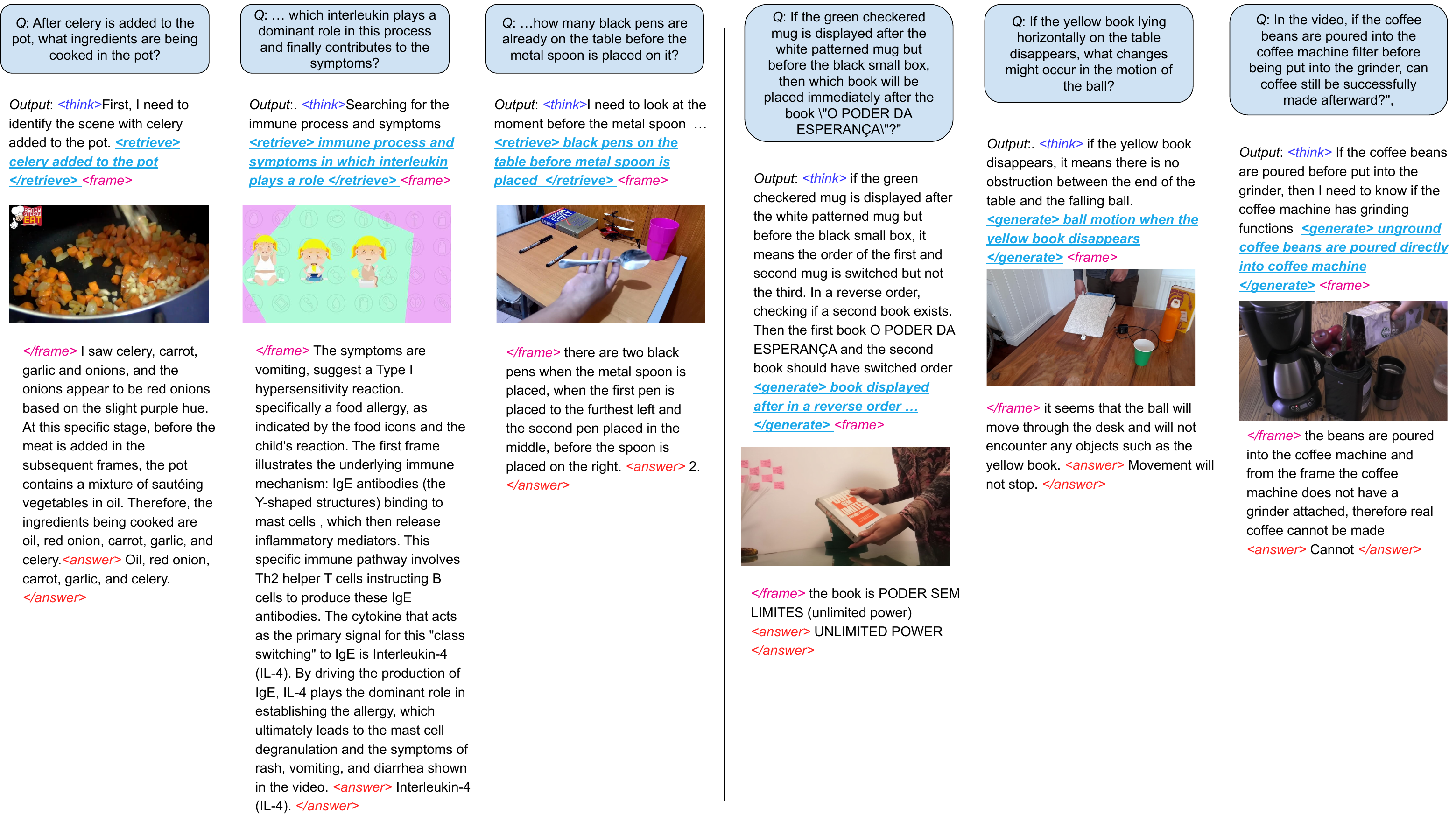}
  \caption{Qualitative examples (full CoT) showing the emergent capability of \Ours outputting retrieved (left half) or generated frames (right) inside the CoTs at inference time, dynamically adding requested visual information (factual original videos via retrieval or counterfactual synthesis via conditional image generation) to the reasoning process.} 
  \label{fig:inference_examples_appendix}
\end{figure*}

\paragraph{FrameMind} \citep{ge2025famemind} is proposed as an end-to-end reinforcement learning framework that facilitates dynamic video reasoning by interleaving textual generation with active visual perception. Through a Frame-Interleaved Chain-of-Thought (FiCOT), the model iteratively detects informational deficits and actively queries targeted visual inputs, such as high-resolution frames, rather than processing static sequences. This adaptive capability is supported by Dynamic Resolution Frame Sampling (DRFS) and optimized via DRFS-GRPO, a group-relative policy optimization algorithm that derives sampling strategies from sparse rewards without frame-level supervision. Ultimately, this methodology enables the flexible balancing of temporal coverage and spatial detail to achieve efficient video understanding.

\paragraph{ViTCoT} \citep{zhang2025vitcot} ViTCoT introduces a Video-Text Interleaved Chain-of-Thought (CoT) paradigm in which a multimodal LLM first generates an initial textual reasoning trace from the original video, question, and options, and then refines this reasoning by interleaving it with a key-video composed of task-relevant frames. The key-video is obtained by automatic key-frame selection with a powerful MLLM and subsequent multi-annotator human verification. This two-stage prompting scheme is model-agnostic and can be plugged into various CoT variants (e.g., standard CoT, Desp-CoT, Plan-and-Solve), yielding more human-like, visually grounded reasoning, better exploitation of critical temporal cues, and richer neuron activation patterns in complex video understanding scenarios. 
\paragraph{Chain-of-Frames} \citep{ghazanfari2025chain} advances video understanding in Multimodal Large Language Models (LLMs) by introducing temporally grounded, step-by-step reasoning. This approach involves fine-tuning video LLMs on COF-DATA, a large-scale dataset of diverse, frame-aware reasoning traces generated efficiently from both real and synthetic videos. By explicitly referencing relevant frames within the reasoning process, CoF integrates temporal information directly into the chain-of-thought structure. This method is self-contained, eliminating the need for the auxiliary networks or complex inference frameworks required by existing approaches. Consequently, CoF enhances model interpretability through explicit temporal grounding and significantly improves performance across various video understanding tasks while reducing hallucinations.
\paragraph{ReWatch-R1} \citep{zhang2025rewatch}  advances complex video reasoning in Large Vision-Language Models through an integrated approach combining agentic data synthesis and process-oriented reinforcement learning. It introduces a multi-stage pipeline to generate the ReWatch dataset, featuring temporally dense captions, challenging multi-hop questions, and video-grounded CoT data. A core innovation is the use of a Multi-Agent ReAct framework for CoT synthesis, which simulates iterative information retrieval and verification against video content. The model is subsequently trained via Supervised Fine-Tuning and Reinforcement Learning with Verifiable Reward (RLVR), incorporating a novel Observation and Reasoning (O$\&$R) reward. This mechanism evaluates both the final answer accuracy and the factual grounding of intermediate reasoning steps, thereby explicitly penalizing hallucinations and enhancing the model's capacity for verifiable temporal reasoning.

\section{Emergent full example}
We include the full CoTs of Figure 4 in the main text in Figure \ref{fig:inference_examples_appendix}. As the examples show, emergent capabilities of actively requesting to retrieve or generate frames is enabled by \Ours, and demonstrating that the reasoning process is enhanced by the additional visual information.

\begin{figure}[th]
  \vspace{-8pt}

	\centering
  \includegraphics[width=0.95\linewidth]{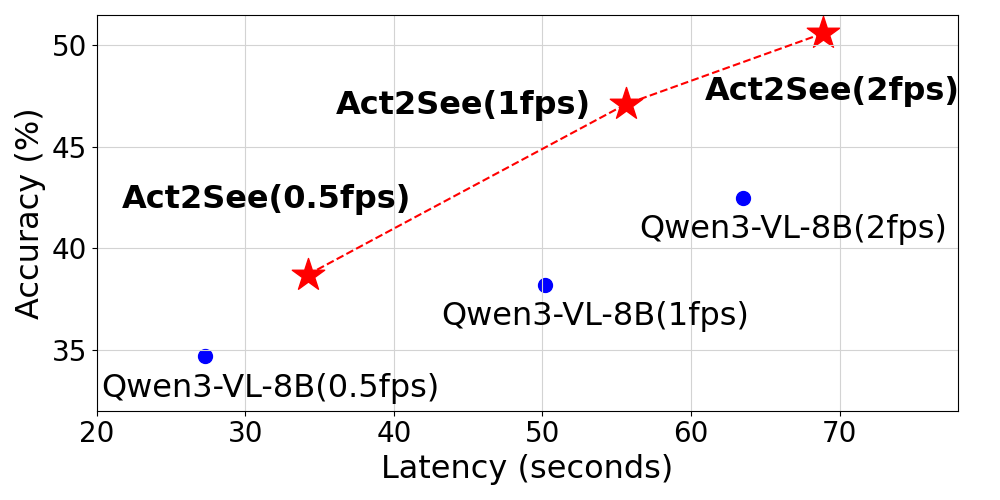}
    \vspace{-10pt}

  \caption{\Ours pushes the Pareto frontier beyond Qwen3-VL-3B baseline by achieving higher accuracy with lower latency.}
  \label{fig:latency_comparison}
  \vspace{-8pt}
\end{figure}

\section{Further Analyses}
Below we include further analyses including latency and FLOPs, when image generation will help reasoning, categorization of image generation failure inside CoTs and their impact, and model's robustness towards generation failure.

\paragraph{Latency and FLOPs.} We report latency (excluding data-loading) and FLOPs (counting all GPU operations, including retrieval and generation) amortized over the full VCR-Bench, and varying the number of frames to the VLMs. Figure \ref{fig:latency_comparison} and Table \ref{tab:flops} show that \Ours achieves higher accuracy, while having lower latency and lower FLOPs than the Qwen3-VL-8B baseline when using half as many frames, demonstrating Pareto dominance.

\begin{table}[th]
\vspace{-8pt}

\caption{\Ours also pushes the Pareto frontier on FLOPs.}
\vspace{-6pt}

\begin{adjustbox}{max width=0.45\textwidth}

\begin{tabular}{cccc}
\toprule
Method & Avg. input frames & Amortized TFLOPs & Accuracy ($\%$)\\
\midrule

Qwen3-VL-8B-Thinking & 1fps (210 frames) & 1025 & 38.2\\
\rowcolor{lightgray} \Ours  & 0.5fps (106 frames) &\textbf{877} & \textbf{38.7} \\
\bottomrule
\end{tabular}
\end{adjustbox}
\label{tab:flops}
\vspace{-6pt}
\end{table}

\paragraph{When will generation help.} We analyze this using counterfactual questions in VCR-Bench. Because VCR-Bench does not provide ground-truth counterfactual labels, we use GPT-5.2 to classify the questions, identifying 107 as counterfactual questions. We then split VCR-Bench into a counterfactual set and a factual set. Table \ref{tab:resutls_by_question_type} shows that: 1) Generation significantly outperforms retrieval-only and no-retrieval-or-generation on counterfactual questions ($29\%$ and $58\%$ relative gains); and 2) Combining retrieval and generation yields the best overall performance.

\begin{table}[th]
\vspace{-6pt}

\caption{Accuracy ($\%$) by question type, showing that generation significantly boosts performance on counterfactual questions, and combining retrieval and generation yields best performance.}
\vspace{-6pt}

\begin{adjustbox}{max width=0.45\textwidth}

\begin{tabular}{ccccc}
\toprule
CoT type & \textbf{Counterfactual set} & \textbf{Factual set}  & \textbf{Full set} \\
\midrule
No-retrieval-or-generation  & 35.52 & 38.51 & 38.20 \\
Retrieval-only & 43.41 & 45.80  & 45.55 \\
Generation-only & 56.01 & 37.70  & 39.60 \\
\rowcolor{lightgray} Retrieval + Generation (\Ours) & \textbf{56.22} & \textbf{46.05} & \textbf{47.10} \\
\bottomrule
\end{tabular}
\end{adjustbox}
\label{tab:resutls_by_question_type}
\vspace{-6pt}
\end{table}

\paragraph{Categorize when generation fails.} Following \citet{borji2023qualitative}, we categorize failures as: geometry (disproportionate size or shape), physics (physically unrealistic), and semantic (wrong spatial relationship, attributes, and composition). Among 127 generation calls on VCR-Bench, we found 13 failures ($10.2\%$): 7 semantic, 4 physics, and 2 geometry. Among the CoTs with the generation failures, semantic failures yield lowest accuracy ($28.6\%$) and predominantly affect counterfactual questions (4 out of 5 cases).

\paragraph{Robustness towards generation failure.} Due to semantic failures being the most common, we study robustness towards generation failure by injecting deliberate semantic failures to visual generation prompts (e.g., word flips like ``red ball''$\to$``yellow ball'') at $10\%$, $20\%$, $30\%$ rates on VCR-Bench. From Table \ref{tab:robustness_towards_generation_failure}, \Ours remains robust and still outperforms retrieval-only at $30\%$ failure rate. We argue \Ours is robust to realistic generation failure as we observe a  $10.2\%$ real-world failure rate the analysis above.

\begin{table}[th]
\vspace{-6pt}

\caption{Accuracy with varying generation failure rates.}
\vspace{-8pt}

\begin{adjustbox}{max width=0.45\textwidth}

\begin{tabular}{ccccc}
\toprule
\textbf{Accuracy ($\%$)} & \textbf{10$\%$ failure} & \textbf{20$\%$ failure}  & \textbf{30$\%$ failure} &  \textbf{Retrieval-only}\\
\midrule
\Ours  & 46.88 & 46.40 & 45.69 & 45.56 \\
\bottomrule
\end{tabular}
\end{adjustbox}
\label{tab:robustness_towards_generation_failure}
\vspace{-8pt}
\end{table}



\end{document}